\documentclass{article} 
\usepackage{iclr2024_conference,times}
\usepackage{graphicx}

\usepackage{amsmath,amsfonts,bm}

\def\eqref#1{equation~\ref{#1}}

\def\1{\bm{1}}

\DeclareMathAlphabet{\mathsfit}{\encodingdefault}{\sfdefault}{m}{sl}
\SetMathAlphabet{\mathsfit}{bold}{\encodingdefault}{\sfdefault}{bx}{n}

\usepackage{hyperref}
\usepackage{url}
\usepackage{float}

\title{Exploring selective image matching methods for zero-shot and few-sample unsupervised domain adaptation of urban canopy prediction}

\author{John Francis \\ 
The Alan Turing Institute\\
London, United Kingdom\\
\texttt{jfrancis@turing.ac.uk} \\
\And
Stephen Law \\
Department of Geography \\
University College London \\
London, United Kingdom \\
\texttt{stephen.law@ucl.ac.uk} \\
}

\iclrfinalcopy 
\begin{document}

\maketitle
\begin{abstract}
We explore simple methods for adapting a trained multi-task UNet which predicts canopy cover and height to a new geographic setting using remotely sensed data without the need of training a domain-adaptive classifier and extensive fine-tuning. Extending previous research, we followed a selective alignment process to identify similar images in the two geographical domains and then tested an array of data-based unsupervised domain adaptation approaches in a zero-shot setting as well as with a small amount of fine-tuning. We find that the selective aligned data-based image matching methods produce promising results in a zero-shot setting, and even more so with a small amount of fine-tuning. These methods outperform both an untransformed baseline and a popular data-based image-to-image translation model. The best performing methods were pixel distribution adaptation and fourier domain adaptation on the canopy cover and height tasks respectively.

\end{abstract}

\section{Background}
Machine learning algorithms utilizing remote sensing data have been able to estimate numerous features of the environment such as building footprints \citep{zhufootprint}, land cover types \citep{valilandcover}, and urban tree canopy characteristics \citep{FRANCIS2023128115}. While these algorithms are often able to produce reliable estimates for a particular geographical area, when images from a new environment, or images captured by different instruments at potentially different resolutions are introduced, the reliability of these algorithms falters \citep{Tuia_2016}. Additionally, as computer vision models are shared more broadly, either through community driven packages (e.g., https://sci.vision/), or are integrated into web tools which allow users to upload their own images for inference (e.g., https://segment-anything.com/demo), unsupervised domain adaptation (\emph{UDA}) methods become critical for increasing the utility of these models to new users with their own unique datasets. 

In this paper, we examine a multi-task UNet model which simultaneously predicts urban canopy cover as well as canopy height from RGB aerial imagery. This algorithm was trained and evaluated solely using images from Chicago, USA, captured by the National Agriculture Imagery Program NAIP. Here we explore methods for applying this algorithm to a Borough in London, UK, using data collected by the UK's Ordnance Survey. Instead of fine-tuning or training a domain-adaptive-classifier, our novelty lies in experimenting a variety of simple data-based \emph{UDA} approaches in a zero-shot setting, which doesn't require any training, or with a small amount of fine-tuning.  

\section{Related works}

Unsupervised domain adaptation (\emph{UDA}) techniques leverage models trained in one source domain to perform accurately on a different, unlabeled target domain. These methods can be broadly categorized into model-based and data-based adaptations.

Model-based adaptations modify the underlying source model to improve its performance on the target domain, primarily through adversarial or self-training methods. \cite{ganin2016domain} introduced a domain-classifier that adjoins to the source model, employing a gradient reversal layer to adjust the model’s weight for the target domain. \cite{tzeng2017adversarial} generalised this notion, by pre-training an encoder on source images and then adapting the target encoder using an adversarial classifier. Self-training, on the other hand, generates target pseudo labels from models trained from the source domains, which are then used to iteratively refine the model for the target domain \citep{zou2018unsupervised}. 

In contrast, data-based adaptation focuses on altering images either in the source or target domain so that the two are statistically similar. This can be accomplished through unpaired image-to-image translation models like CycleGAN \citep{zhu2017unpaired}, which suffers from changes in semantic content. More recent efforts such as ColorMapGAN \citep{tasar2020colormapgan} try to tackle this limitation by restricting the model to only perform pixel-wise intensity transformations, achieving promising results. However, both model-based approaches and image-to-image data-based approaches require a separate model to be learned between domain pairs, a challenge (energy and costs considerations) that becomes impractical with an increasing number of domains \cite{strubell2019energy}. 

Instead, this research focuses on more straightforward data-based approaches with simple image-matching methods, whereby no training or modifications are made to the underlying source model or the need for an image translation model. Related works in remote sensing includes \cite{yaras2023randomized}, who designed a random-selective histogram matching algorithm for searching and then aligning similar images from source and target domain. Expanding upon this randomised selective search, we evaluate an array of image matching algorithms beyond histogram matching 
for tree canopy cover and height prediction. 

\section{Method}

Our aim in this research is to test the efficacy of selective-search image-matching methods to align an RGB image $I\in\mathbb{R}^{hxwx3}$ in the new target domain (London) to be used on models trained in the source domain (Chicago) in a zero-shot setting or without extensive fine-tuning and data augmentation. In this work, we transform an image $I_{tgt}$ in the target domain into the same or similar image $I_{tgt \rightarrow src}$ in the source domain by $I_{tgt \rightarrow src} = T_{tgt \rightarrow src}(I_{tgt})$ so that a model trained in the source domain $Y_{src}=f_{src}(I_{src};\theta)$ parameterised by $\theta$ can be used for prediction using the translated target image $Y_{tgt \rightarrow src}=f_{src}(I_{tgt \rightarrow src})$ without retraining or extensive fine-tuning or data augmentation. 

\begin{figure}[H]
    \centering
    \includegraphics[width=0.75\textwidth]{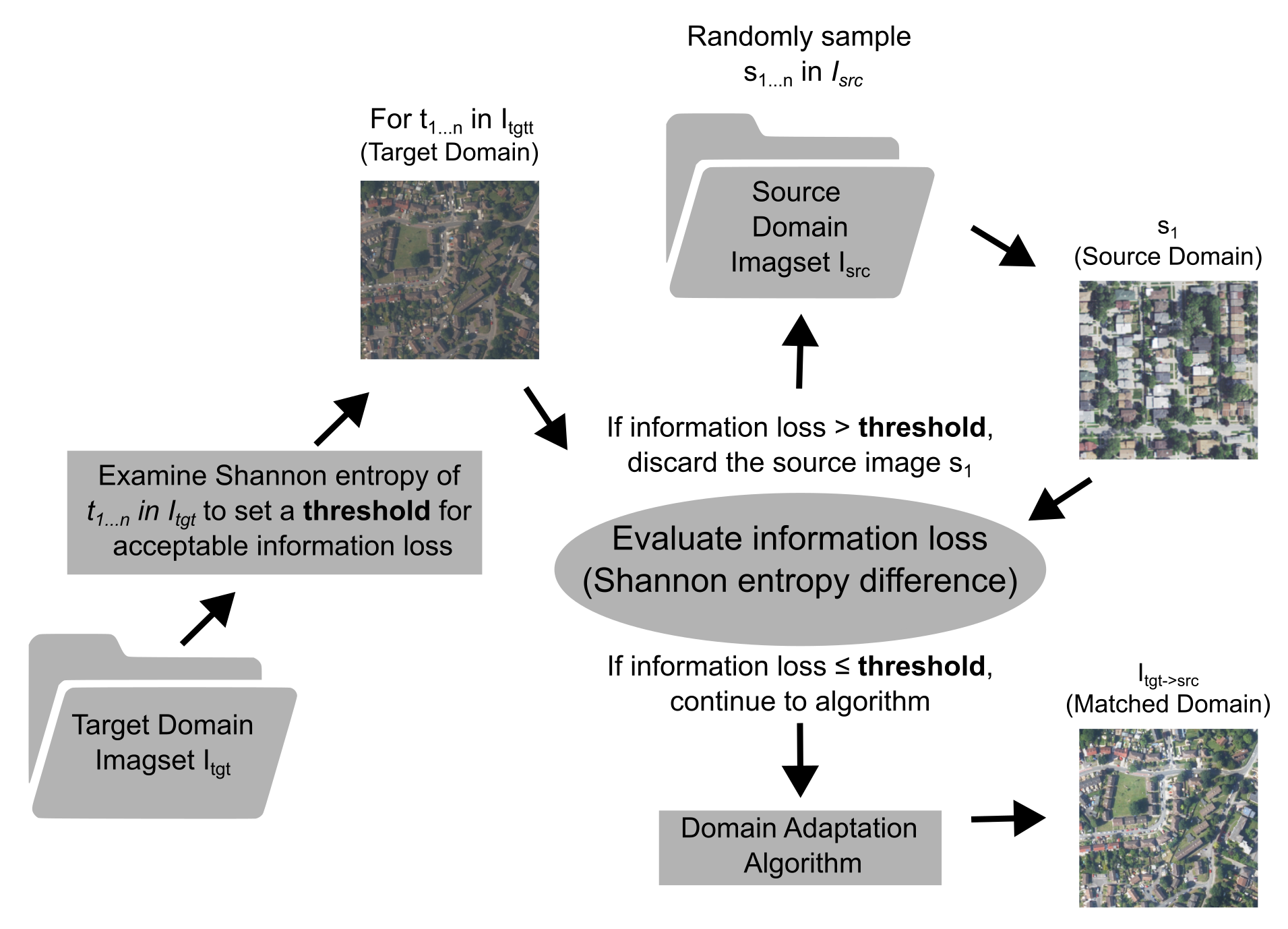}    \caption{\label{fig:matching}Randomized Image Matching}
\end{figure}

The data for this study includes 812 1m resolution RGB images from each of London, UK \footnote{\url{https://digimap.edina.ac.uk/aerial}}and Chicago, USA \footnote{\url{https://naip-usdaonline.hub.arcgis.com/}} covering about 47 square kilometres. London images were chosen based on their alignment with available and temporally aligned LiDAR point cloud data at roughly 12 points per square metre \footnote{\url{https://environment.data.gov.uk/survey}} which could be used to generate estimates of ground truth tree canopy cover and height for evaluation, using methods similar to \cite{ROUSSEL2020112061, FRANCIS2023128115}. Chicago images were randomly selected to provide a comparable set of source domain images for the domain adaptation algorithms. The London images were originally collected at a 25cm resolution while the Chicago images were collected at a 1m resolution, so prior to any domain adaptation approaches the London images were converted to 1m using a simple mean aggregation.

Figure \ref{fig:matching} shows the randomized image matching process we used to align images from London and Chicago. Similar to \cite{yaras2023randomized}, we used the Shannon entropy measure $S = -\sum p(x)log(p(x))$ to evaluate the likelihood of a paired reference image leading to a large amount of information loss during the translation of an image from the new target domain to the source domain. We randomly sampled images from the source domain (Chicago) until a viable match was found to ensure minimal information loss in the images of the target domain (London), leading to more robust transformations.

We tested four simple data-based image matching approaches, histogram matching in RGB space \citep{yaras2023randomized}, histogram matching in LAB space, fourier domain adaptation (FDA) \citep{yang2020fda}\footnote{\url{https://github.com/YanchaoYang/FDA}} and a naive pixel distribution adaptation (PDA) that simple transforms(PCA) an image fitted in the source domain and then inverse transform fitted from the target domain. The transformation were completed using the scikit-image \citep{scikit-image} and Albumentations \citep{info11020125} python packages. 

We additionally tested one data-based image-to-image translation model, CycleGAN \cite{zhu2017unpaired}, as well as the target (London) images without transformation, as baselines. We evaluate each of these methods in a zero-shot setting, testing the pre-trained UNet \citep{FRANCIS2023128115} in the source domain of Chicago on the transformed images, and then fine-tune separate versions of the algorithm using the transformed datasets. The canopy cover task is evaluated using intersection over union (IoU), while the canopy height task is evaluated using mean absolute error (MAE).

For model fine-tuning, the 812 transformed images in the target domain were split into a train and test set using an 80/20 split. The algorithm was tuned for a maximum of 75 epochs with early stopping if the model failed to improve after 10 epochs. An Adam optimizer with an initial learning rate of 0.001 was used, decaying by a gamma of 0.905 every five epochs. Following \cite{FRANCIS2023128115}, the loss function is a simple weighted additive loss which is a combination of mean squared loss for the canopy height task and binary cross entropy for the tree canopy cover task $L_{total}=\gamma L_{canopy} + \lambda L_{height}$ where we chose the best parameter $\lambda$ based on previous work in the source domain. 

\section{Results}

\begin{figure}[ht]
    \centering
    \includegraphics[width=0.95\textwidth]{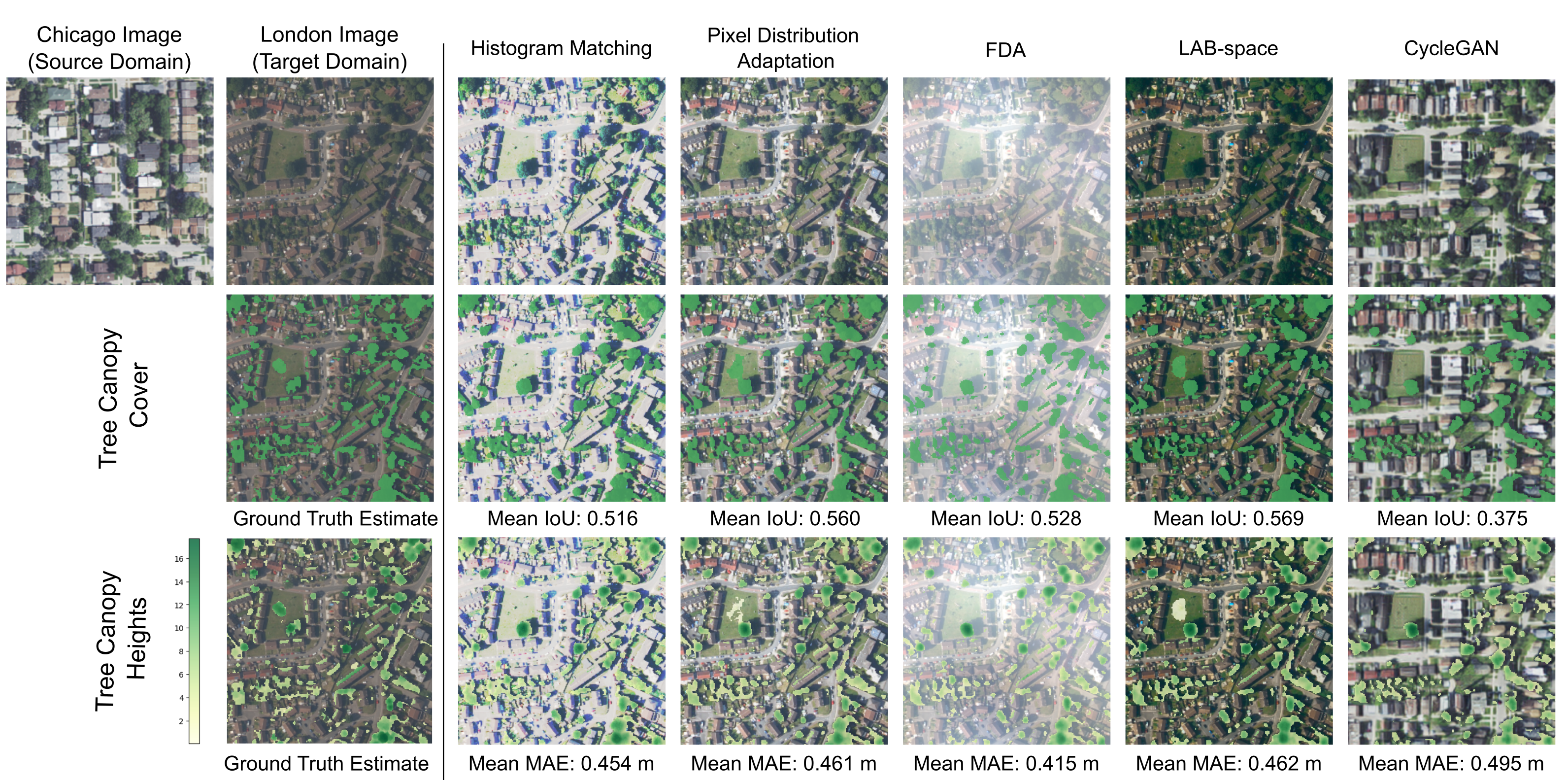}    \caption{\label{fig:uda_methods}Example outputs from different Unsupervised Domain Adaptation Methods.}
\end{figure}

Figure \ref{fig:uda_methods} shows an example image transformation for each of the methods evaluated in this paper, alongside its ground truth and predicted estimates for tree canopy cover and height. All the data adaptation methods have perceptually altered the pixel intensity of the target image. PDA in particular made the target image slightly brighter and more similar to the source domain. Conversely, CycleGAN has shifted the target image closest to the source domain but the semantics of the image have also been unrealistically altered (e.g. changes in buildings and roads) \citep{tasar2020colormapgan}.

Table \ref{tab:zeroshot} shows the zero-shot results for the source model evaluated on the transformed target images. For the canopy cover task, PDA produced the best results with a mean IoU of 0.5131. FDA and LAB-space histogram matching both performed worse than evaluating the original target images without any image transformations. Conversely, for the canopy height task, the most performative method was FDA with a MAE of only 0.5864, which is even better than the results of the algorithm on the source domain images. Histogram matching, PDA and CycleGAN all performed worse on the canopy height task than the original target images without any image transformations.

The results of the small samples fine-tuned algorithms can be seen in Table \ref{tab:finetune}. Similar to the zero-shot results, PDA was the best-performing method for the canopy cover task with an IoU of 0.7014. Histogram matching, LAB-space histogram matching, and CycleGAN all performed worse than a model fine-tuned on the un-transformed target images. For the canopy height task, PDA also had the lowest MAE at 0.5547 narrowly edging past LAB-space histogram matching. All of the simple image matching methods outperformed the algorithm fine-tuned on the un-transformed target images for the canopy height task, while CycleGAN performed worse than the un-transformed images on both tasks. 

\begin{table}[ht]
  \centering
  \begin{minipage}[b]{0.45\linewidth}
    \caption{Zero-shot Unsupervised Domain Adaptation}
    \label{tab:zeroshot} 
    \begin{tabular}{|c|c|c|c|}
    \hline
        Data & Method & mIoU & MAE(m)  \\ \hline
        Source & NA & 0.6650 & 0.6538  \\ \hline \hline
        Target & None & 0.3709 & 0.8321  \\ \hline
        Target & HM & \underline{0.4784} & 1.0302  \\ \hline
        Target & FDA & 0.3659 & \textbf{0.5864}  \\ \hline
        Target & PDA & \textbf{0.5131} & 0.9312  \\ \hline
        Target & LAB-HM & 0.3695 & \underline{0.6062}  \\ \hline
        Target & Cyclegan & 0.4010 & 1.2085  \\ \hline
    \end{tabular}
  \end{minipage}
  \quad 
  \begin{minipage}[b]{0.45\linewidth}
    \centering
    \caption{Unsupervised Domain Adaptation w/Fine-Tuning(small samples)} 
    \label{tab:finetune} 
    \begin{tabular}{|c|c|c|c|}
    \hline
        Data & Method & mIoU & MAE(m)  \\ \hline
        Source & NA & 0.6650 & 0.6538  \\ \hline \hline
        Target & None & 0.6885 & 0.6340  \\ \hline
        Target & HM & 0.6811 & 0.5944  \\ \hline
        Target & FDA & \underline{0.6917} & 0.5839  \\ \hline
        Target & PDA & \textbf{0.7014} & \textbf{0.5547}  \\ \hline
        Target & LAB-HM & 0.6862 & \underline{0.5627}  \\ \hline
        Target & Cyclegan & 0.5714 & 0.8519  \\ \hline
    \end{tabular}
  \end{minipage}
\end{table}

\section{Discussion}
The selective-aligned data-based image transformation approach is able to outperform simple benchmarks using translated remotely sensed images from London to generate predictions with a model trained on Chicago for a multi-task urban canopy cover and height prediction algorithm in both zero-shot and few-sample fine-tune settings. Fine-tuning consistently outperforms zero-shot methods in our tests, however we still find the zero-shot methods useful as it may not always be practical to fine-tune an algorithm for every new target domain introduced. 

Despite being able to align the distribution of London's image pixels to be more similar to the Chicago images the algorithm was originally trained on, these methods cannot fully address the domain shift. This can arise from differences in geographical features, such as house shape, neighborhood layout, and building density specific to our target domain, London, which were not present in the original image set, Chicago, as well as discrepancies in the conditions and instruments that were used to capture the imagery. Because of this, there seems to be a limit on how well these methods can perform zero-shot domain adaptation. In many scenarios however, there may not be viable ground truth estimates with which to fine-tune an algorithm on, or the algorithm may be operating within a very low-resource secure environment, such as a data-safe-haven, where any additional model training would be extremely time consuming. In these situations, FDA and PDA offered the best performance on canopy height and canopy cover predictions respectively, outperforming the baseline methods without the need to access the underlying model.  

Surprisingly, nearly all of the fine-tuned algorithms outperformed the results the original algorithm acquired on the original image set. It is possible that by combining information from multiple locales, the total predictive power contained within the algorithm was increased. However, we did not have enough data to train a model from scratch on the London image set to compare against the fine-tuned model performance. Another possibility is that London is simply an easier setting for the task of canopy cover and height prediction than Chicago, as the buildings are much shorter on average.

While we tested these domain adaptation methods on an algorithm shifting from Chicago to London, future work would benefit from exploring domain shifts across a wider variety of geographic locales. Future work could also explore more advanced model-agnostic methods to improve the results in a zero-shot setting, test the methods over not only a one-to-one scenario but also many-to-many geographical scenarios, incorporate location information into the adaptation \citep{marsocci2023geomultitasknet} or adaptation with large scale satellite image foundation models \cite{bastani2023satlaspretrain}.

\section{Conclusion}
Overall, we find that selective-aligned simple image matching approaches can be an effective way to adapt a multi-task deep learning algorithm to a new geographic setting, especially when a small amount of fine-tuning is possible. Fourier domain adaptation and pixel distribution adaptation obtained the best results on our two tasks of canopy cover and canopy height prediction, although simple histogram matching techniques still provided moderate boosts over not using a domain adaptation method. Simple data-based domain adaptation approaches such as these provide easy, low-resource methods to enhance the utility of pre-trained models which utilize remote sensing imagery, when these algorithms are exposed to new, unseen domains. In many scenarios, fine-tuning or completely retraining an algorithm whenever it is exposed to a new domain can be either difficult to implement or prohibitively expensive. Researchers which are attempting to utilize pre-trained remote sensing algorithms should explore simple image matching techniques, before extensive and costly retraining, when applying these models to new geographic settings.

\section{Acknowledgements}
This work was supported by Towards Turing 2.0 under the EPSRC Grant EP/W037211/1, the Ecosystem Leadership Award under the EPSRC Grant EP/X03870X/1 \& The Alan Turing Institute.

\bibliography{iclr2024_conference}
\bibliographystyle{iclr2024_conference}


\end{document}